\documentclass[runningheads]{llncs}

\usepackage{eccv}

\usepackage{eccvabbrv}

\usepackage{graphicx}
\usepackage{booktabs}
\usepackage{makecell}
\usepackage{bbding}
\usepackage{multirow}

\usepackage[accsupp]{axessibility}  % Improves PDF readability for those with disabilities.

\usepackage{hyperref}

\usepackage{orcidlink}

\begin{document}

\title{Combining Generative and Geometry Priors for Wide-Angle Portrait Correction} 

\titlerunning{Generative and Geometry Priors}

\author{Lan Yao\inst{1} \and
Chaofeng Chen\inst{2} \and
Xiaoming Li\inst{1}\,\Envelope \and
Zifei Yan\inst{1} \and
Wangmeng Zuo\inst{1,3}}

\authorrunning{Lan~Yao et al.}

\institute{Harbin Institute of Technology \and
Nanyang Technological University \and
Pazhou Lab, Huangpu\\
\email{\{mrha011010,chaofenghust,csxmli\}@gmail.com  \{yanzifei,wmzuo\}@hit.edu.cn}
}

\maketitle

\begin{abstract}
Wide-angle lens distortion in portrait photography presents a significant challenge for capturing photo-realistic and aesthetically pleasing images. Such distortions are especially noticeable in facial regions.
In this work, 
we propose encapsulating the \emph{generative face prior} as a guided natural manifold to facilitate the correction of facial regions.
Moreover,
a notable central symmetry relationship exists in the non-face background, yet it has not been explored in the correction process.
This \emph{geometry prior} motivates us to introduce a novel constraint to explicitly enforce symmetry throughout the correction process, thereby contributing to a more visually appealing and natural correction in the non-face region. 
Experiments demonstrate that our approach outperforms previous methods by a large margin, 
excelling not only in quantitative measures such as line straightness and shape consistency metrics but also in terms of perceptual visual quality.
All the code and models are available at \url{https://github.com/Dev-Mrha/DualPriorsCorrection}.

\end{abstract}

\section{Introduction}
\label{sec:intro}

Due to the convenience of taking pictures with mobile phones and the advantages of wider viewing angles, people frequently choose the wide-angle mode for photographing individuals. However, the distortion caused by wide-angle lenses often results in images with distorted lines and stretched faces, greatly affecting the aesthetic quality of portrait photography.

\begin{figure}[tb]
  \centering
  \includegraphics[width=\linewidth]{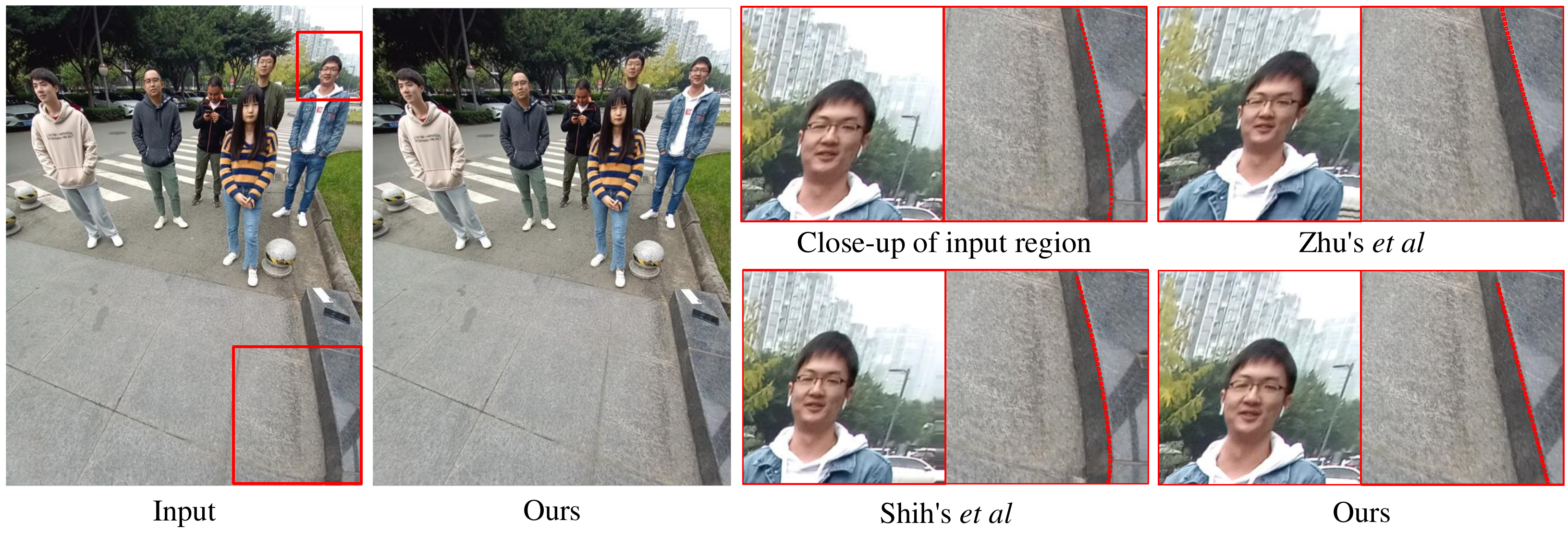}
  \caption{Example of distorted and corrected images in comparison to other methods. Thanks to the proposed generative and geometry priors, our results have straighter lines in the background and more natural-looking faces compared to other methods.}
  \label{fig:intro}
\end{figure}

Traditional methods focus on optimizing a warping map to correct distortions from different projections, such as perspective, Mercator, and stereographic projections \cite{2009Optimizing,2010Image,2018A}. 
Since human faces are usually quite close to the camera and are severely affected by these distortions, it cannot simply be performed simultaneously with background correction. Therefore, previous works usually handle face and background regions using different methods. Shih \etal \cite{shih2019distortion} proposed a solution by combining perspective projection for the background and stereographic projection for the face region, using a warp mesh. 
They approached it as an optimization problem, seeking an optimal balance between redressing the background and the faces. 
Recent approaches \cite{tan2021practical,zhu2022semi} introduced deep learning methods to tackle this issue, eliminating the need for additional parameters. 
Although great progress has been achieved, the real-world faces may suffer from various types of distortions that are challenging to correct without additional information, especially when training data is limited (see Fig.~\ref{fig:intro}).  

In this paper, we introduce a framework that leverages generative and geometry priors to rectify wide-angle distortions in faces and backgrounds, respectively.
We follow previous practice \cite{tan2021practical} and separate our framework into two modules: FaceCNet, which targets facial distortion, and LineCNet, which addresses the straightness of buildings and lines in the background. The FaceCNet employs the generative face structure prior from pre-trained StyleGAN to enhance facial correction. 
Our key motivation is that the pre-trained StyleGAN is able to map the latent vectors $\mathcal{W}$ to a normally shaped face through GAN inversion \cite{Abdal_2019_ICCV,tov2021designing}. Given that $\mathcal{W}$ is a compact latent space,  it facilitates the mapping of distorted faces into $\mathcal{W}$ thereby restoring facial structures.
This structural guidance enables FaceCNet to effectively correct wide-angle facial distortions with complex projection. 
For the background regions, we empirically observe that they mainly suffer from barrel distortion and exhibit central symmetry. 
We therefore introduce symmetric regularization in the training of LineCNet.
This simple geometric prior proves effective in straightening lines in the background.
Finally, we design a post-processing step to harmoniously fuse the corrected face region into the background.
The experimental results show our method produces undistorted images with well-proportioned faces, surpassing existing techniques in both visual quality and quantitative evaluation metrics.
The main contributions are:
\begin{itemize}
\item[$\bullet$] We integrated the generative face prior from StyleGAN to aid in correcting wide-angle distortions of facial regions. This face structure prior enables our method to handle faces distorted by complex projections.
\item[$\bullet$] 
To the best of our knowledge, we are the first to employ a symmetry prior in the background of wide-angle photos, which is highly effective in correcting curved lines.
\end{itemize}

\section{Related Works}

\subsection{Wide-angle Portrait Correction}

Traditional approaches to correcting wide-angle lens distortion often rely on mathematical formulations and optimization problems, identifying specific formulas to mitigate the distortion effect \cite{zorin1995correction,zhang1999flexible,Valente_2015_CVPR_Workshops,carroll2009optimizing}. However, due to the complexity of real-world problems that often resist ideal mathematical modeling, recent methods have increasingly turned to deep learning-based approaches. Zhao \etal \cite{zhao2019learning} utilized camera calibration to obtain parameters for predicting flow with U-Net, facilitating fisheye undistortion. Shih \etal \cite{shih2019distortion} employed the field of view (FOV) to compute a mesh, integrating various distortions in the face region and background. Cao \etal \cite{caoadaptive} introduced an adaptive triangle mesh for wide-angle portrait correction. However, these methods require additional camera parameters, which are often not available. Therefore, recent methods have attempted to directly learn to correct the distortion. Tan \etal \cite{tan2021practical} pioneered the use of deep learning methods to remove distortions without relying on camera parameters, employing a two-stage network. Zhang \etal \cite{Zhang_2023_CVPR} proposed a method for wide-angle correction through content-aware conformal mapping, applying different distortion recovery models to different parts. Zhu \etal \cite{zhu2022semi} presented a semi-supervised transform-based network to exploit semantic information for portrait correction. Unlike these approaches, our method employs geometry and generative priors for background and facial regions, respectively, achieving better results for real-world portraits.

\subsection{Face Prior as Reference}
Structure priors have been proven effective in many low-level vision
tasks, \eg, depth image enhancement~\cite{li2016deep,gu2017learning}, image inpainting~\cite{dolhansky2018eye,nazeri2019edgeconnect,ren2019structureflow,li2020learning}, and image restoration~\cite{li2018learning,li2020blind,li2020enhanced,chen2021progressive,li2022learning,wang2021towards,chan2021glean,yang2021gan,chen2022real,jiang2021robust}.
With the rapid advancement of generative adversarial networks (GANs), particularly exemplified by StyleGAN \cite{Karras_2020_CVPR}, researchers have explored the potential of pre-trained GAN models to provide comprehensive priors, including both structural and textural attributes for many low-level tasks~\cite{wang2021towards,chan2021glean,yang2021gan,li2023learning}. Numerous GAN inversion methods \cite{richardson2021encoding,tov2021designing,alaluf2021restyle,gu2020image} have emerged, enabling the manipulation of real images by optimizing their latent codes within the GAN space~\cite{liu2023survey}. 
This capability has also led to the adoption of StyleGAN for various face restoration tasks \cite{pan2021exploiting,menon2020pulse,chan2021glean,yang2021gan}. For example, GFP-GAN \cite{wang2021towards} integrates a degradation module with a pre-trained StyleGAN on facial data, achieving remarkable performance in various face restoration tasks, such as super-resolution, inpainting, colorization \etc. The primary advantage of employing generative priors lies in their capability to effectively generate missing regions within an image. 
This generative power holds particular promise for correcting wide-angle distortions in facial images, as missing regions pose a challenge to straightforward correction via direct warping techniques. 
%In this work, we make the first attempt to embed the generative face prior to the wide-angle portrait correction task.

\section{Method}

\subsection{Architecture Overview}

\begin{figure}[tb]
    \centering
    \includegraphics[width=1\textwidth]{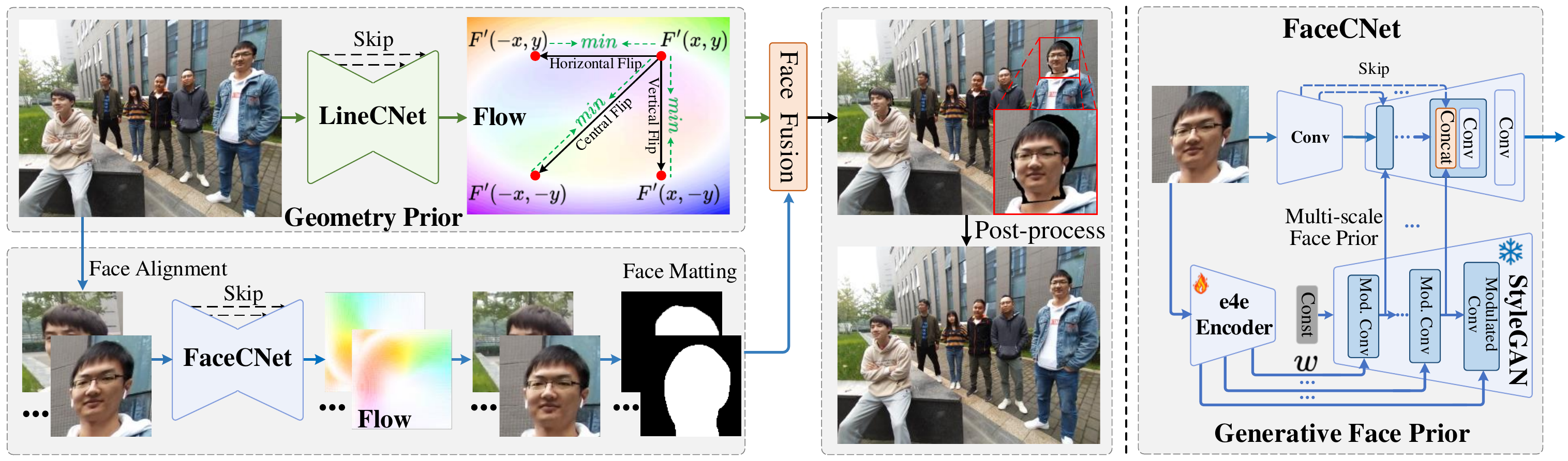}
    \caption{
    {Overview of our framework. It contains three parts: LineCNet for the geometric distortions in the background, FaceCNet for correcting the face region using the generative face prior, and a post-process step for fusing the corrected face region into the background.}}
    \label{fig:architecture}
\end{figure}

{
Our approach is capable of correcting wide-angle portrait that preserves the natural appearance of both the face and background regions. 
As illustrated in \cref{fig:architecture}, the whole framework is divided into three parts: LineCNet, FaceCNet, and face fusion block. 
Given a wide-angle image, 
LineCNet is applied to address geometric distortions affecting straight lines in the background while FaceCNet is introduced to correct the face regions with the guidance of generative face prior from StyleGAN.
The face fusion block finally merges the undistorted faces generated by FaceCNet with the background corrected by LineCNet. 
}

\subsection{Generative Prior for Face Correction}
In comparison with other objects, the face region owns specific structures that are easily perceptive when exhibiting unnatural distortions. Therefore, in the correction process, we should concentrate more on the facial region. However, existing methods mainly consider facial landmarks as a constraint, overlooking overall facial consistency and naturalness. Notably,
the generative models (\eg, StyleGAN) have encapsulated the ability to generate photo-realistic face images, enabling us to facilitate face correction towards a more natural performance. 
In the following, we first review StyleGAN inversion for real-world images and then introduce the method of encapsulating the generative prior for face correction.

The original StyleGAN models generate face images from random Gaussian noise. To adapt StyleGAN models to real-world images, previous works map the face image into the $\mathcal{W}$ space. 
This inversion process is to predict the latent representation of an image generated by a Generative Adversarial Network (GAN), aiming to find the latent code that, when input into the GAN generator, reproduces the target image. 
In our approach, we follow these methods and employ the encoder for editing (e4e) framework \cite{tov2021designing} for StyleGAN inversion.
The e4e method \cite{tov2021designing} revolves around training an encoder network to map real images to their corresponding latent codes within the $\mathcal{W}+$ latent space of StyleGAN2, which is a high-dimensional approximation of StyleGAN’s $\mathcal{W}$ space. Once trained, this encoder network can deduce the latent code of an input image by locating the closest code in the learned latent space that would produce a similar image when processed by the StyleGAN2 generator. The inferred latent code enables various manipulations such as facial expression alterations, age progression/regression, and other image modifications. The e4e encoder provides a precise means for image editing by directly manipulating the latent space of StyleGAN2.

\begin{figure}[t]
\centering
\includegraphics[width=\linewidth]{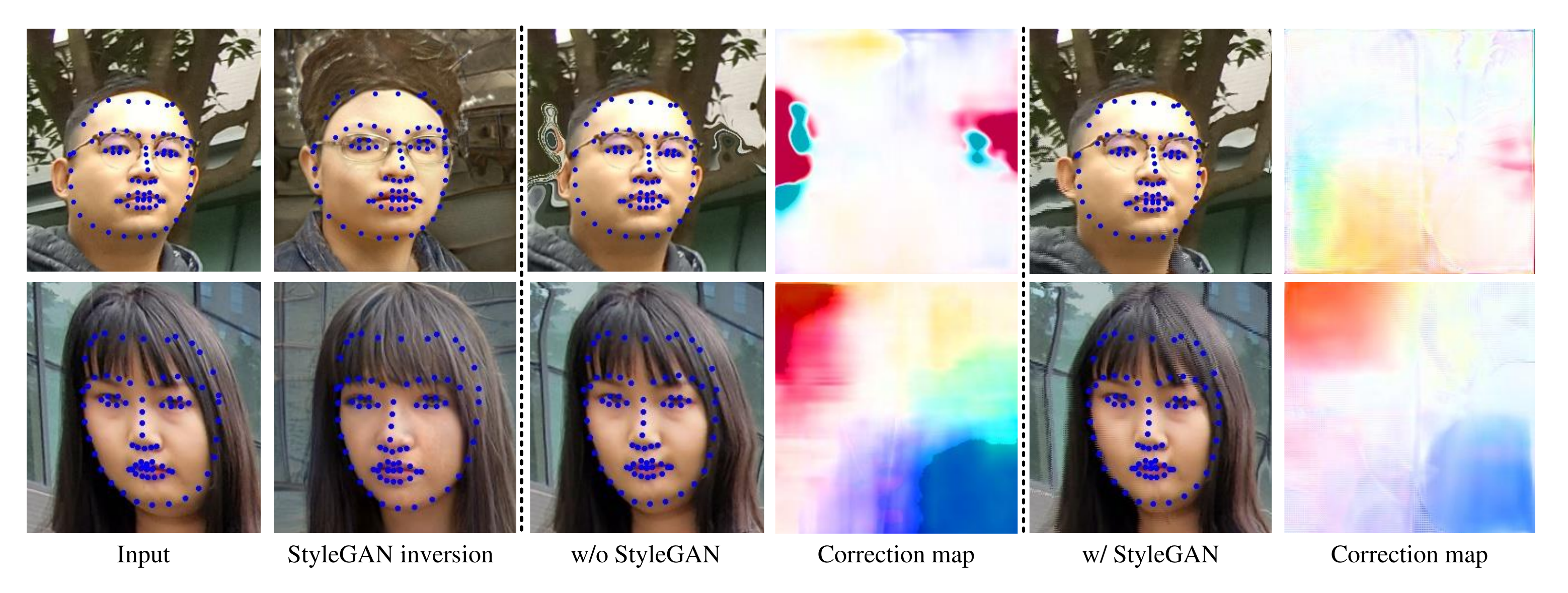}
\caption{Examples of our face correction w/ and w/o using generative prior. Through the results of correcting flow, it can be observed that the correction is concentrated at the position where the StyleGAN face structure deviates from the input.}
\label{fig:face}
\end{figure}

During the inversion process, it is inevitable to lose facial structures, resulting in lower fidelity. This implies that directly applying StyleGAN to correct the face region fails to preserve identity textures related to facial identity (refer to the StyleGAN inversion results in \cref{fig:face}). Since the intermediate features in StyleGAN can generate photo-realistic structures that align well with the natural manifold, we propose using these priors to guide the correction process. Specifically, for face images, we first obtain their $\mathcal{W}$ vector using the pre-trained e4e encoder. We then utilize the intermediate features of StyleGAN to aid in the prediction of deformation flow, preserving higher fidelity. Additionally, the e4e encoder is pre-trained on high-quality face images. To generalize to unnatural faces, we fine-tune the encoder with our synthetic distorted faces exhibiting characteristics similar to wide-angle distortion stretching. Here, we adopt a synthetic optical flow similar to the distortion produced by a wide-angle lens to generate distorted faces for training our FaceCNet. Leveraging the fine-tuned e4e model\cite{tov2021designing}, distorted faces are inverted into the StyleGAN domain, facilitating the correction of distorted face images.

\cref{fig:architecture} (right) demonstrates the details of our FaceCNet. 
It predicts the face correction flow with the input of faces and their corresponding StyleGAN features. 
To achieve this, we employ the U-Net architecture~\cite{ronneberger2015u} as a baseline, chosen for its ability to calculate cost volumes for flow prediction. 
The design of U-Net enables the fusion of features from various receptive fields, facilitating comprehensive information processing.

Moreover, we introduce multi-scale feature fusion into the U-Net framework to integrate StyleGAN features in different feature spaces.
We analyze that the incorporation of multi-scale feature fusion serves a dual purpose. Firstly, it enables the seamless integration of StyleGAN features, leveraging the rich priors associated with facial structures. This integration is crucial for an effective correction process, as it allows FaceCNet to better understand and preserve intricate details inherent to facial images. Secondly, the multi-scale feature fusion contributes to the overall robustness of the network by accommodating information from different scales, ensuring that FaceCNet can capture both fine-grained and global features in the correction task. 
By incorporating StyleGAN features, FaceCNet gains access to additional information that goes beyond the standard input faces. This additional information proves instrumental in preserving natural facial details and enhancing the overall quality of corrected images. Consequently, the described architecture and feature integration approach contribute to FaceCNet's ability to produce natural-looking results in the facial regions.

\subsection{Symmetry Prior for Background Correction}

%\begin{figure}[t]
%\centering
%\includegraphics[width=.98\textwidth]{./figs/wide-angle.pdf}
%\caption{Illustration of symmetry deformation in the wide-angle images. }
%\label{fig:symmetry}
%\end{figure}

Wide-angle lenses typically produce images with barrel distortion, which is a specific type of distortion related to the lens itself.
Incorporating the radial distortion figure highlights the lens-induced nature of radial distortion, further supporting the argument for symmetry within barrel distortion.
%, as demonstrated in \cref{fig:symmetry}.
Barrel distortion arises as a geometric distortion due to the optical design of the lens. This occurs because the outer regions of the lens magnify the image more than the central portion, leading to a noticeable bulging effect towards the edges of the frame.
Barrel distortion follows a specific format, and it is related to camera parameters\cite{zhang1999flexible,fitzgibbon2001simultaneous}, as described by \cref{eq: barrel}:
\begin{equation}
    r_{u} = r_{d} \times (1 - k_1 \times r_{d}^2 - k_2 \times r_{d}^4)\,,
    \label{eq: barrel}
\end{equation}
where $r_{d}$ represents the distance from the center of distortion in the distorted image, $r_{u}$ represents the undistorted distance, and $k_1$ and $k_2$ are coefficients related to the lens's distortion characteristics. 
These coefficients are determined during camera calibration and are used to correct barrel distortion in image processing workflows.
From this formula and comparison chart, it can be found that barrel distortion is only related to the camera lens, and increases as the distance from the center grows.

\begin{figure}[t]
  \centering
  \begin{minipage}{0.28\linewidth}
    \includegraphics[width=\linewidth]{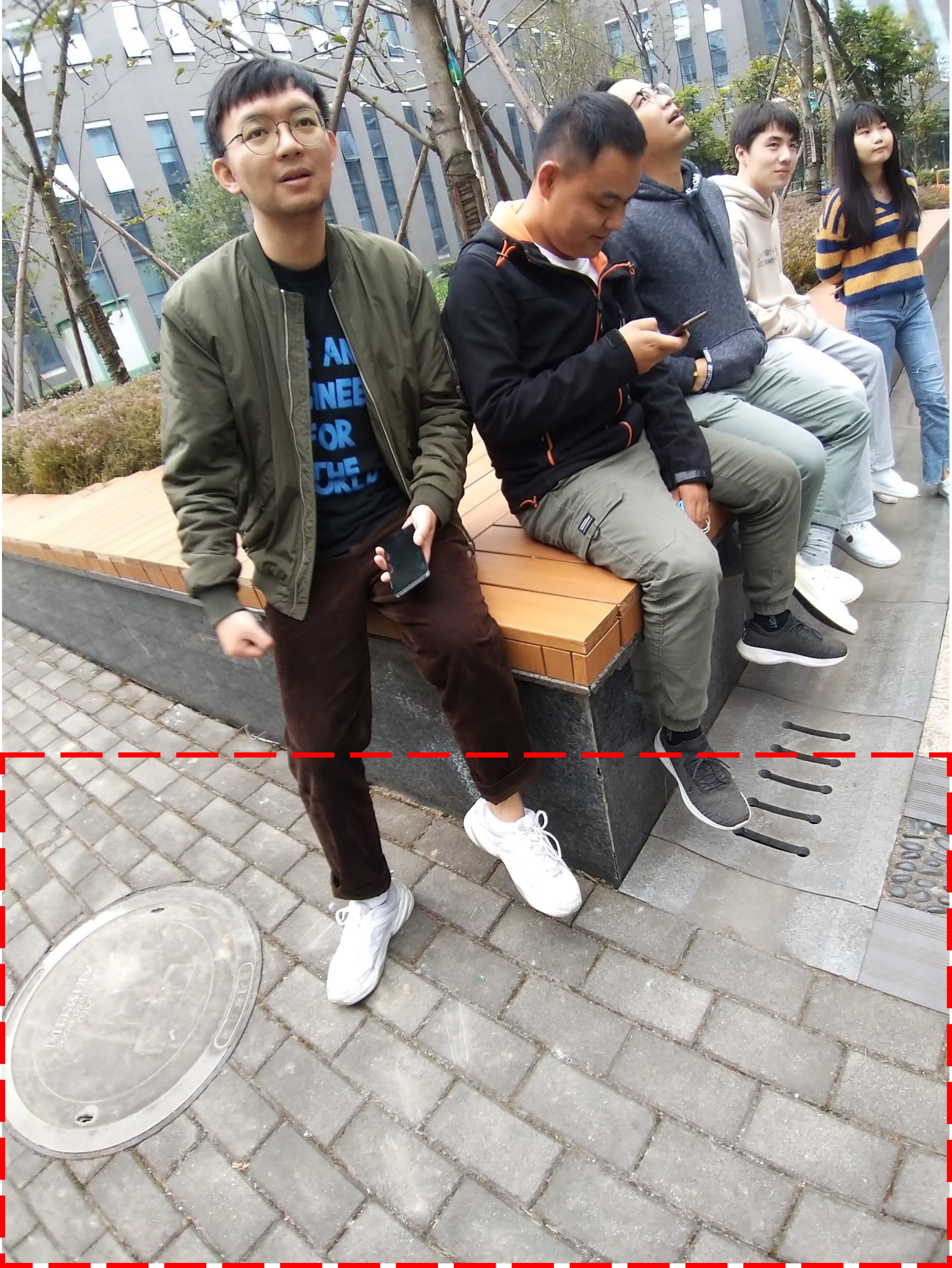}
  \end{minipage}
  \begin{minipage}{0.6\linewidth}
  \begin{subfigure}{0.48\linewidth}
    \includegraphics[width=\linewidth]{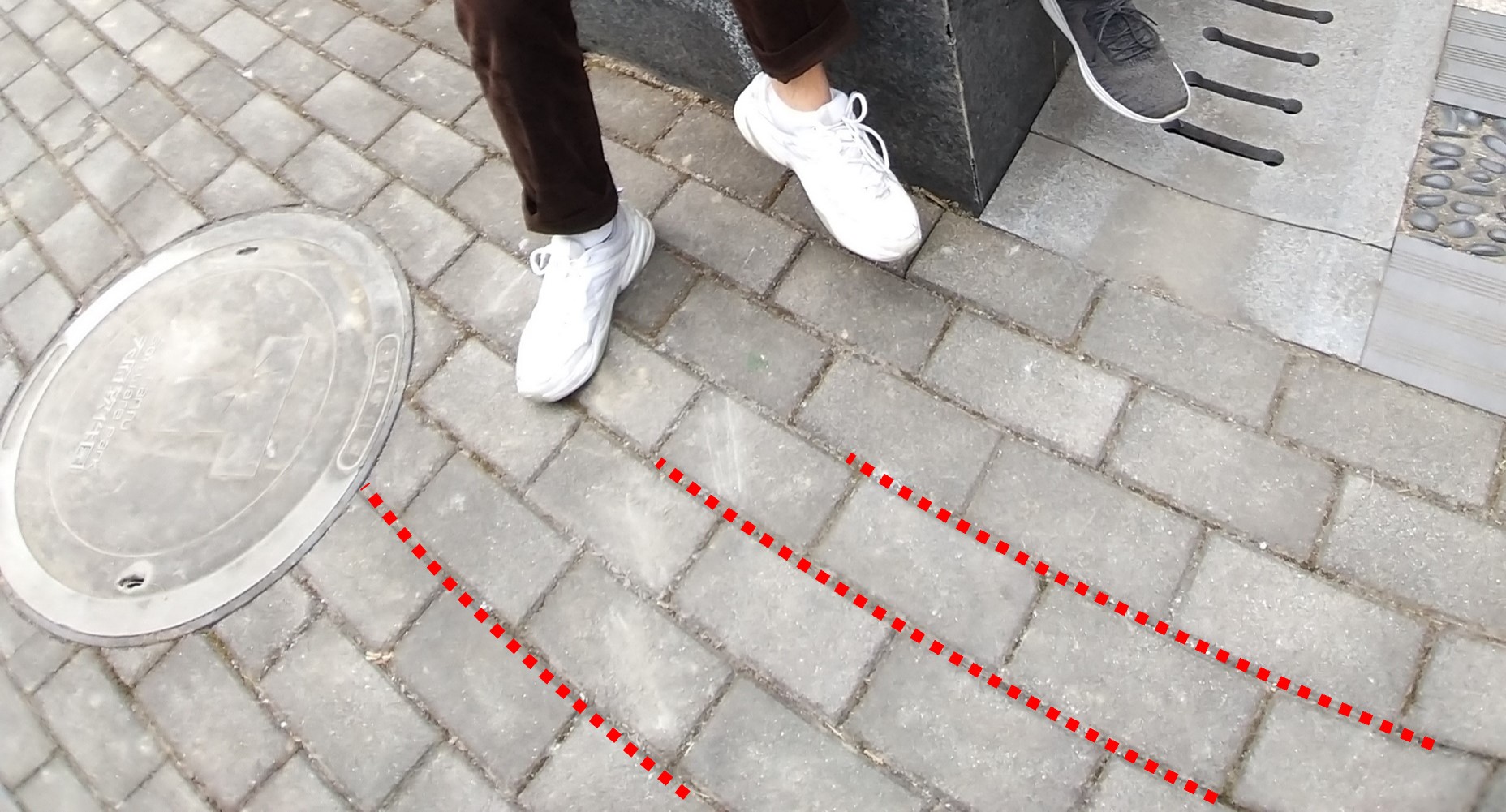} 
    \caption{Origin}
  \end{subfigure}
  \begin{subfigure}{0.48\linewidth}
    \includegraphics[width=\linewidth]{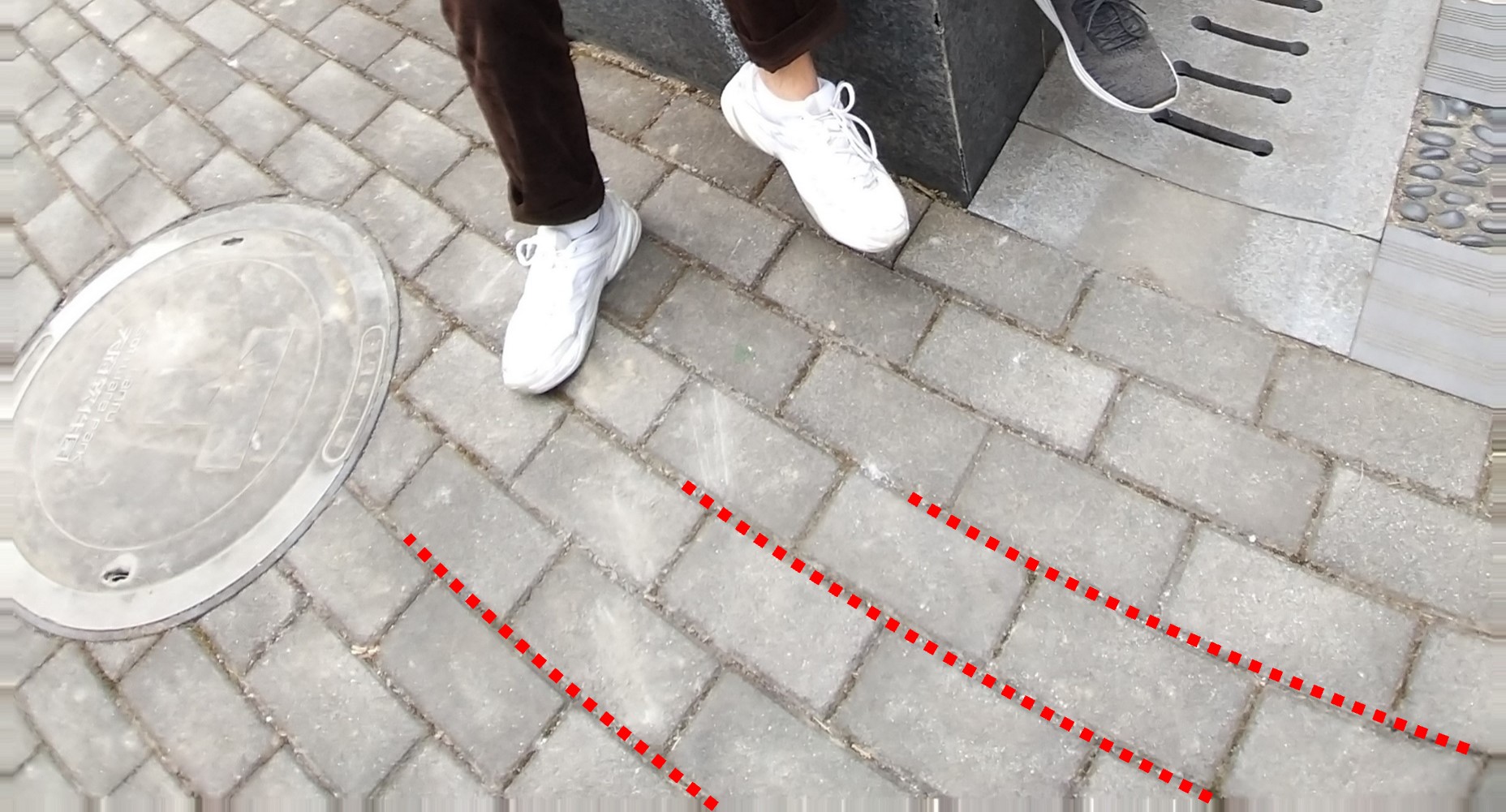} 
    \caption{Shih's \etal~\cite{shih2019distortion}}
  \end{subfigure}
  
  \begin{subfigure}{0.48\linewidth}
    \includegraphics[width=\linewidth]{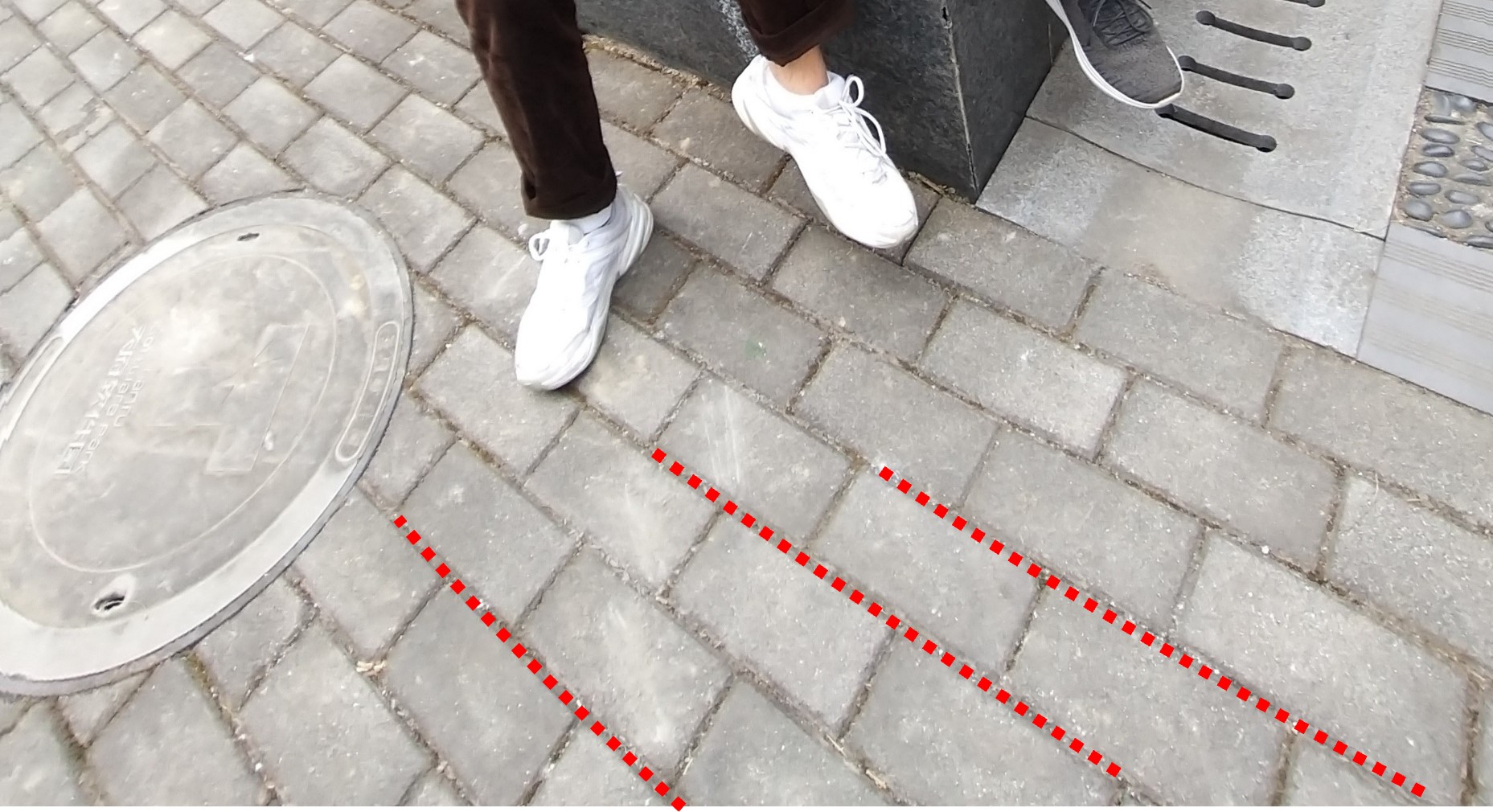} 
    \caption{Zhu's \etal~\cite{zhu2022semi}}
  \end{subfigure}
  \begin{subfigure}{0.48\linewidth}
    \includegraphics[width=\linewidth]{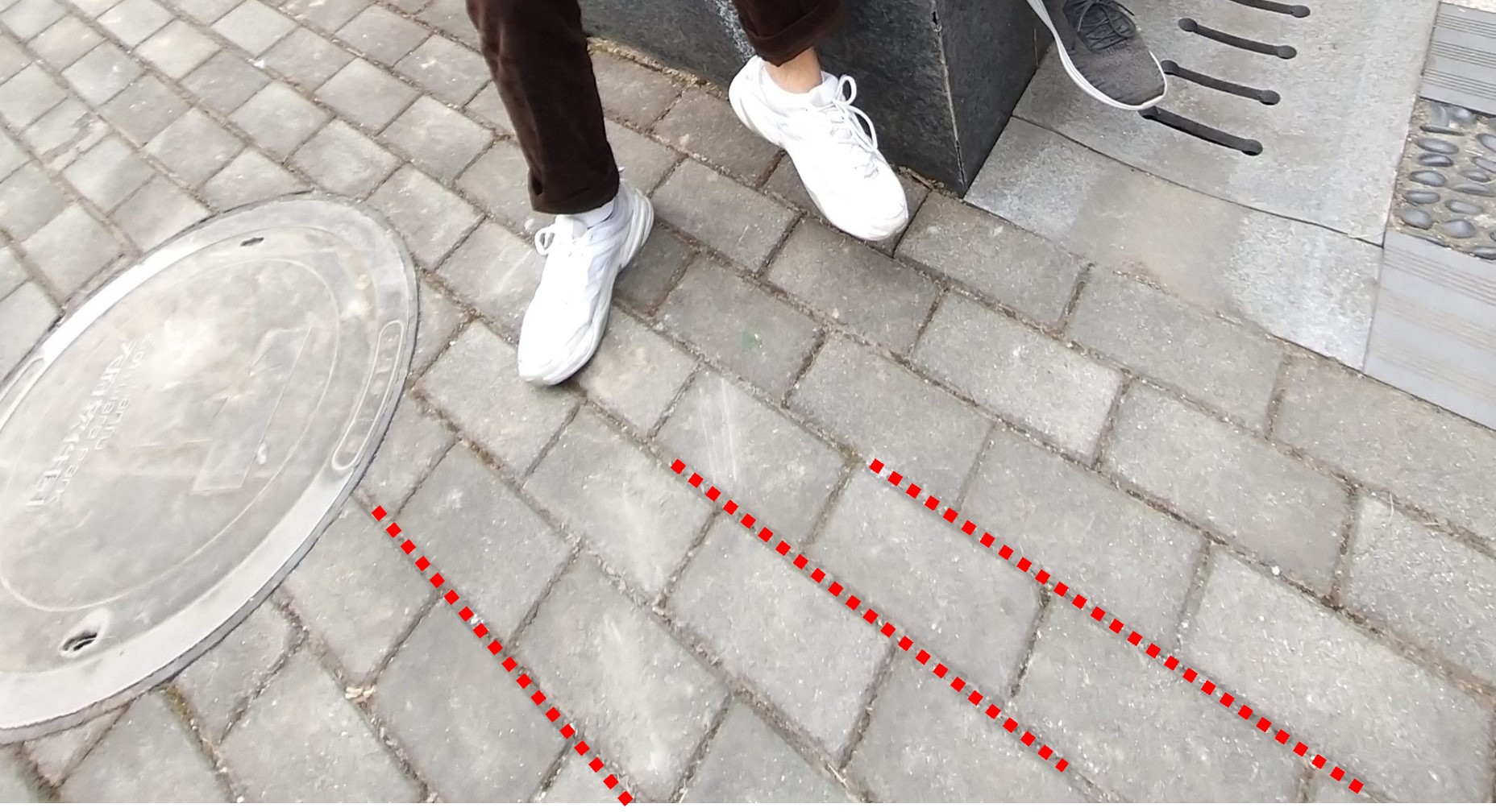} 
    \caption{Ours}
  \end{subfigure}
  \end{minipage}
  \caption{Comparison results of background correction with others. We can observe that the proposed LineCNet can better correct the distorted lines.}
  \label{fig:sym_result}
\end{figure}

Indeed, perspective distortion is another aspect of visual distortion commonly encountered in wide-angle photography
, but it differs from barrel distortion in its origin and characteristics \cite{8960275,SANTANACEDRES20171}. 
Unlike barrel distortion, which is a geometric distortion caused by the optical properties of the lens, perspective distortion is an undesired effect in real images that is not caused by lens flaws.
Perspective distortion is related to the relative sizes and distances of objects within the scene, as well as the positioning and angle of view of the camera.
Objects close to the lens appear abnormally large relative to more distant objects, and distant objects appear abnormally small and hence farther away.
When perspective distortion interacts with stereo objects, it can disrupt the accurate alignment of features between images, complicating depth perception and 3D reconstruction in stereo imaging applications. 
That's why we cannot use common undistortion methods to correct stereo objects like human faces.

Therefore, without considering spherical objects such as human faces, the distortion introduced by wide-angle lenses can be considered symmetrical.
From this, we introduce symmetric loss to help correct the optical flow training of the background part.
According to the optical flow to recover distortion should have horizontal, vertical, and central symmetry, a symmetric regularization loss is designed and applied in our LineCNet to effectively correct the distortion and reduce the complexity of the network without predicting the camera parameters.

In particular, our LineCNet, similar to FaceCNet, utilizes the U-Net network as its baseline architecture. 
As mentioned above, the distortion caused by the wide-angle lens to the image is consistent with symmetry, so no other modules except symmetry loss are added to the LineCNet.
This is driven by the aim of maintaining simplicity and efficiency in addressing background distortions. 
By leveraging the U-Net and symmetry constraint, LineCNet can effectively correct the background distortions in wide-angle images as shown in \cref{fig:sym_result}.

\subsection{Face Fusion and Postprocess} 

In contrast to previous approaches, we adopt a novel strategy of isolating the face from the background and subsequently reassembling the faces during the merging process.
This strategy ensures that the straight lines surrounding the face in the background remain unaffected. 
Therefore, our FaceCNet can correct the face region towards a more photo-realistic performance while struggling to have pleasing results on the other region (see background in~\cref{fig:face}). 
To address this issue, we employ face parsing on cropped facial regions, allowing only the precise facial components to be inserted. 
ParseNet \cite{chen2021progressive} is utilized for this purpose which separated the face and background in the crop face image and obtained the mask of the face part.
When reattaching the face to the image, we calculate the translation position of the face before and after LineCNet correction to ensure a more accurate fit between the face and the corrected background. 
In this process, face correction and background correction will cause a certain gap in the region, especially the region near the border.
Image in-painting algorithm Lama \cite{suvorov2021resolution} or other diffusion-based methods~\cite{lugmayr2022repaint} is finally used to fill this missing region. 

This comprehensive approach ensures that both facial and background components are corrected separately and then integrated seamlessly, leading to high-quality, and distortion-free images.

\subsection{Learning Objectives}

Since our FaceCNet and LineCNet are designed for different regions, one for the face and the other for the background, they can be trained in parallel. 

As for training FaceCNet, it is natural to require the predicted flow vector $\Phi$ close to the ground-truth $\Phi^\textit{gt}$, and we thus define the face deformation loss as:
\begin{equation}
    \mathcal{L}_\textit{face}^{f} = \|\Phi_\textit{face} - \Phi^\textit{gt}_\textit{face}\|_{2}^2\,.
\end{equation}
Meanwhile, we also need to ensure that the warped face image $I_\textit{face}^w$ using $\Phi$ approaches to the ground-truth image $I_\textit{face}^\textit{gt}$ in the pixel domain:
\begin{equation}
    \mathcal{L}_\textit{face}^{p} = \|I_\textit{face}^w - I_\textit{face}^\textit{gt}\|_{2}^2\,,
\end{equation}
where $I_\textit{face}^w$ is obtained using the flow interpolation:
\begin{equation}
    \!I_{\textit{face}}^w(i, j)\!=\!\!\!\sum_{(h, w) \in \mathcal{N}} \!I_\textit{face}(h, w) \max\left(0,1-\left|\Phi_{i, j}^y\!-\!h\right|\right) \max \left(0,1\!-\!\left|\Phi_{i, j}^x-w\right|\right)\,.
\end{equation}
$\mathcal{N}$ denotes the 4-pixel neighbors and $I_\textit{face}$ is the face region cropped from the wide-angle portraits.

Finally, we add total variation (TV) loss as smoothness regularization to avoid too large changes in adjacent areas:
\begin{equation}
    \mathcal{L}_\textit{face}^\textit{tv} = \sum_{i,j} \sqrt{(\Phi_{i,j-1} - \Phi_{i,j})^2 + (\Phi_{i+1,j} - \Phi_{i,j})^2}\,.
\end{equation}

To sum up, the final learning objective for FaceCNet is defined as:
\begin{equation}
    \mathcal{L}_\textit{FaceCNet} = \mathcal{L}_\textit{face}^f + \lambda_1\mathcal{L}_\textit{face}^p + \lambda_2 \mathcal{L}_\textit{face}^\textit{tv}\,,
\end{equation}
where $\lambda_1$ and $\lambda_2$ are set to 2 and 0.5, respectively.

In terms of optimizing LineCNet, the loss function is defined as:\begin{equation}
\mathcal{L}_\textit{LineCNet} = \|\Phi_\textit{bg} - \Phi^\textit{gt}_\textit{bg}\|_{2}^2 + \lambda_3 \|I_\textit{bg}^w - I_\textit{bg}^\textit{gt}\|_{2}^2  + \lambda_4 \mathcal{L}_\textit{Sym} \,,
\end{equation}
where $\Phi_\textit{bg}$ and $I_\textit{bg}^w$ are the predicted background flow field and warped background image. $\Phi_\textit{bg}^\textit{gt}$ and $I_\textit{bg}^{\textit{gt}}$ denote their corresponding ground-truth. The symmetry loss $\mathcal{L}_\textit{Sym}$ is defined as:
\begin{equation}
\mathcal{L}_{Sym} = \|\Phi^{v}_\textit{bg} - \Phi_\textit{bg}\|_{2}^2 + \|\Phi^{h}_\textit{bg} - \Phi_\textit{bg}\|_{2}^2  + \|\Phi^{c}_\textit{bg} - \Phi_\textit{bg}\|_{2}^2\,,
\end{equation}
where $\Phi^{v}_\textit{bg}$, $\Phi^h_\textit{bg}$, and $\Phi^{c}_\textit{bg}$ present the vertical, horizontal, and central flip of predicted flow field $\Phi_\textit{bg}$, respectively (see flow map illustration in \cref{fig:architecture}).
This loss function reinforces symmetrical alignment, enhancing the accuracy of line correction. In our experiment, $\lambda_3$ and $\lambda_4$ are set to 1 and 2, respectively.

\section{Experiments}
\subsection{Data Preparation}

\subsubsection{Face dataset}
\begin{figure}[t]
    \centering
    \includegraphics[width=\linewidth]{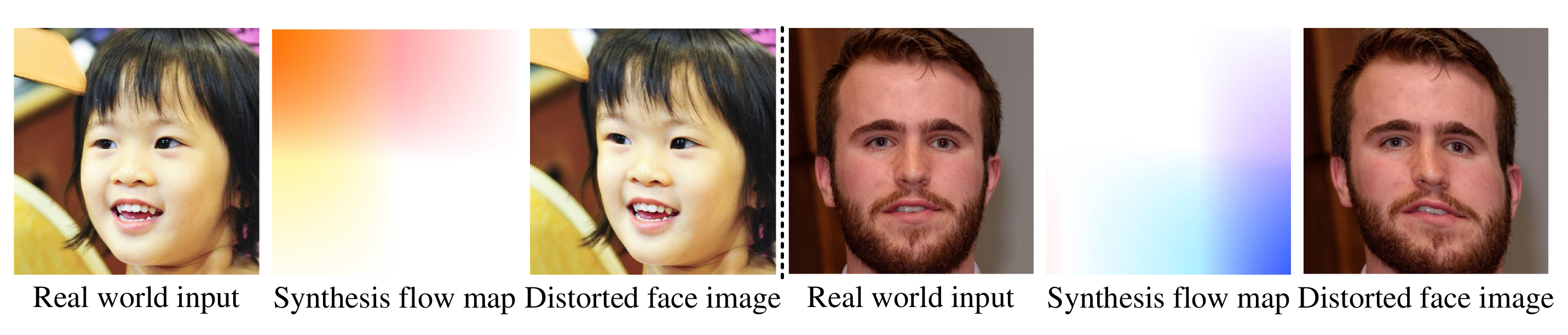}
    
    \caption{Example of our synthetic face pairs for training FaceCNet.}
    \label{fig:facedata}
\end{figure}

The GAN inversion network is trained and evaluated with the public dataset, \ie, FFHQ \cite{karras2019style} and CelebA-HQ \cite{karras2018progressive}.
To generate image pairs for training, we designed a flow that stretches the four corners outward to simulate the distortion caused by a wide-angle lens (see the examples in~\cref{fig:facedata}). 
%We use the FFHQ-1024 dataset for training and CelebA-HQ dataset for validation. 

\subsubsection{Wide angle portrait dataset} We use the public wide-angle portrait dataset from Tan et al. \cite{tan2021practical}. This dataset consists of over 5000 images for training and 129 images for testing. Each item in the dataset contains the ground truth image, the line-corrected image, and the correction flow maps for training. It also provides landmark points of line correction and face correction for quantitative evaluation.

\subsection{Implementation Details}

As shown in \cref{fig:architecture}, each network of our framework is trained separately and the results are combined with a fusion postprocess block.
%Both LineCNet and FaceCNet process images as input and produce correction flows as output. 
%The corrected image is obtained by applying the flow transformation to the input image.
The FaceCNet is trained with the synthetic faces first and then finetuned with real images cropped from \cite{tan2021practical}. The face images are resized to $256 \times 256$ as inputs to FaceCNet.
LineCNet is trained with $512 \times 384$ inputs which is consistent with previous works \cite{tan2021practical,zhu2022semi}. Both of the networks are trained with Adam optimizer \cite{kingma2014adam} and the learning rate is set to $1e-4$. The batch size is $4$ and it takes 5 days to train FaceCNet and 3 days for LineCNet on a 3060 NVIDIA GPU.  
%In order to ensure a fair comparison, all images show the results of the method obtained without any additional calibration pre-processing.

\subsection{Evaluation Metrics}

We follow previous works \cite{tan2021practical} and employ several evaluation metrics including LineAcc and ShapeAcc to assess the performance of our method.

\subsubsection{LineAcc} is designed to evaluate the straightness of lines within the image by comparing the slope variation between the output and ground truth images:
\begin{align}
   LS = 1 - \frac{1}{n} \sum_{i=0}^{n-1} (\frac{y_{d_{i}} - y_{d_{i-1}}}{x_{d_{i}} - x_{d_{i-1}}} - \frac{y_{g_{o}} - y_{g_{n}}}{x_{g_{o}} - x_{g_{n}}}) 
\end{align}
where $LS$ is the similarity between the slopes of lines in the output and ground truth images. $n$ is the number of points in each line. $(x_{g_{i}}, y_{g_{i}})$ and $(x_{d_{i}}, y_{d_{i}})$ are the corresponding positions in the reference and distortion images, respectively.
\subsubsection{ShapeAcc} focuses on evaluating the similarity of facial landmarks between the output and reference images. 
This metric calculates the cosine similarity between corresponding landmark vectors, which is defined as follows:
\begin{align}
    &FC = \frac{1}{n} \sum_{i=0}^{n-1} \cos{(L_{g_{i}}, L_{d_{i}})} 
\end{align}
where $L_{g_{i}}$ and $L_{d_{i}}$ denote the corresponding face landmark points in the reference and distortion images, respectively.

\subsubsection{Landmark Distance} To provide a more comprehensive measurement of facial correction, we introduce the Landmark Distance metric, which calculates the similarity of faces based on the Euclidean distance between corresponding landmark points as below:
\begin{align}
    LD = \frac{1}{n} \sum_{i=0}^{n-1} \|L_{g_{i}} - L_{d_{i}}\|_2^2
\end{align}
For a fair comparison, we align the position of the nose position in reference and distorted images before calculating the distance.

\subsection{Ablation Study}

\begin{table}[t]
\caption{Ablations on the proposed generative and geometry priors.}
\label{table:propose}
\centering
\renewcommand{\arraystretch}{1.1}
  \setlength{\tabcolsep}{1.2mm}
    {
    \begin{tabular}{lccccc}
    \toprule
    \textbf{Method} & \makecell[c]{Symmetry\\ Prior} & \makecell[c]{Generative\\Prior} & LineAcc & ShapeAcc & Landmark Distance\\
    \midrule
    Baseline  &   &   & 66.192 & 97.027 & 5.991  \\
    Baseline + Sym.  & \checkmark &  & 67.304 & 97.266 & 5.5464  \\ 
    Baseline + Gen.  &   & \checkmark & 66.192  & 99.012 & 5.3397 \\
    Ours (\textit{Full})  & \checkmark & \checkmark  & \textbf{67.304} & \textbf{99.012} & \textbf{5.013} \\
    \bottomrule
    \end{tabular}
}
\end{table}

\begin{figure}[t]
    \centering
    \includegraphics[width=\linewidth]{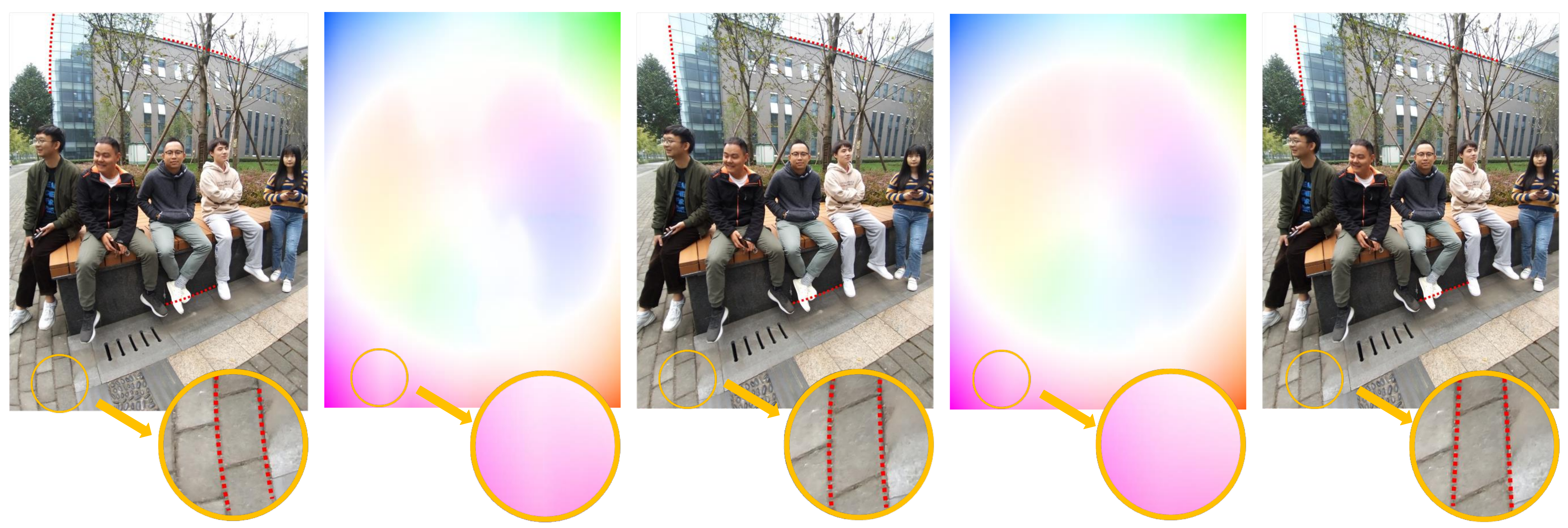}
    \scriptsize{
    \makebox[0.19\linewidth]{(a) Input}
    \makebox[0.39\linewidth]{(b) w/o SymLoss}
    \makebox[0.39\linewidth]{(c) w/ SymLoss}
    }
    \caption{Visual comparison of our method with and without geometric symmetry prior.}
    \label{fig:sym}
\end{figure}

To evaluate the performance of our proposed framework, we first conduct ablation experiments to analyze the effectiveness of symmetry prior and face prior. Then, we conduct experiments to show the importance of a multi-scale feature fusion strategy for face prior. 

\subsubsection{Effectiveness of Priors}
We explore how the priors affect the network performance. 
We utilize the network without any priors as the baseline, and evaluate the performance of four networks with different prior configurations as follows:
\begin{itemize}
    \item[1)] Baseline: LineCNet and FaceCNet with simple U-Net structure and no priors;
    \item[2)] Baseline + Sym. (Symmetry Prior): add symmetry loss for LineCNet;
    \item[3)] Baseline + Gen. (Generative Prior): add generative face prior features from StyleGAN into FaceCNet;
    \item[4)] Ours (\textit{Full}): add both symmetric prior and generative prior.
\end{itemize}

\begin{figure}[!t]
    \centering
    \includegraphics[width=\linewidth]{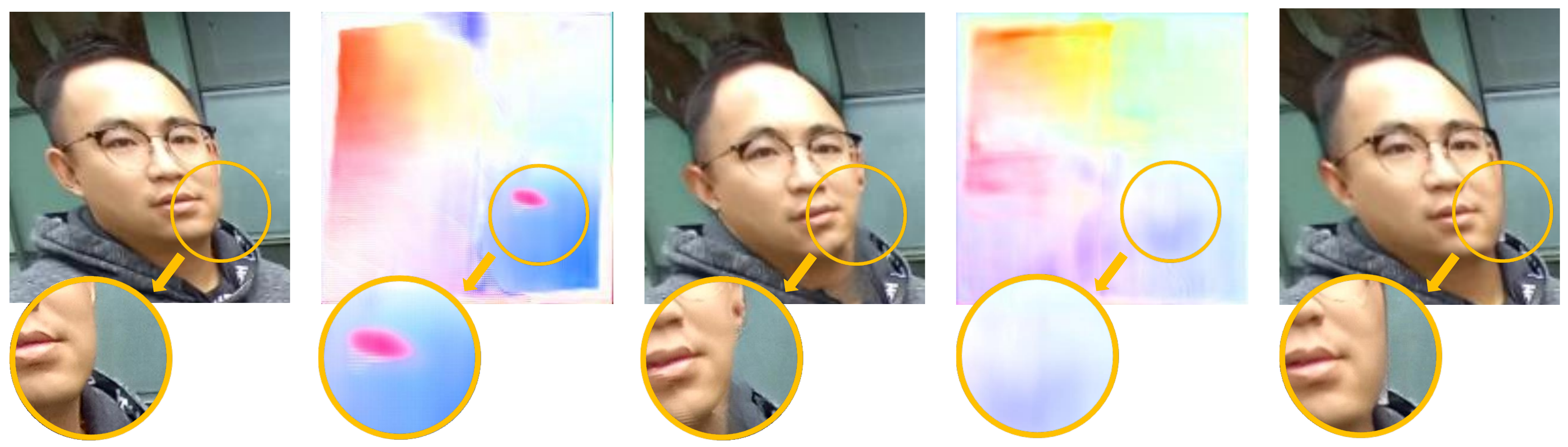}
    \scriptsize{
    \makebox[0.19\linewidth]{(a) Input}
    \makebox[0.39\linewidth]{(b) w/o StyleGAN prior}
    \makebox[0.39\linewidth]{(c) w/ StyleGAN prior}
    }
    \caption{Visual comparison of our method with and without generative face prior.}
    \label{fig:gen}
\end{figure}

The results of the quantitative comparison are provided in \cref{table:propose}. We can observe that the symmetric prior improves the LineAcc while the generative prior benefits the ShapeAcc and Landmark Distance. The effectiveness of them is more obvious from the visual examples in \cref{fig:sym} and \cref{fig:gen}. 

In \cref{fig:sym}, the comparison shows that the predicted flow with $L_{Sym}$ in \cref{fig:sym}~(b) is smoother and more symmetric than that without $L_{Sym}$ in \cref{fig:sym}~(c). From the left bottom corner, we can see that with a more symmetric flow, the contours of bricks in \cref{fig:sym}~(c) are much straighter compared to those in \cref{fig:sym}~(b). The results clearly demonstrate the superiority of using symmetric prior.

\begin{table}[t]
\caption{Ablations on the StyleGAN feature concatenation}
\label{table:ablation1}
\centering
\renewcommand{\arraystretch}{1.1}
  \setlength{\tabcolsep}{5.2mm}
    {
    \begin{tabular}{@{}lccc@{}}
    \toprule
    \textbf{Method} & {PSNR} & {SSIM} & {Landmark Distance}\\
    \midrule
    {1)} Baseline & 25.586 & 0.848 & 3.091 \\
    {2)} Baseline + Single Scale & 24.335 & 0.797 & 2.908 \\
    {3)} Baseline + Multi-Scale & \textbf{26.732} & \textbf{0.871} & \textbf{2.357} \\
    \bottomrule
    \end{tabular}
}
\end{table}

In \cref{fig:gen}, the benefit of employing the StyleGAN prior is more evident. Without the StyleGAN prior, the network predicts an unusual flow near the cheek in \cref{fig:gen}~(b), leading to an odd face shape due to excessive shrinking. Meanwhile, the result in \cref{fig:gen}~(c) appears more natural. This demonstrates that with the guidance of the StyleGAN prior, the network is less likely to produce odd results and is more effective at restoring natural-looking facial structures.

\subsubsection{Multiscale facial prior features}

To evaluate the impact of incorporating multiscale StyleGAN features on face correction, we use a U-Net architecture as our baseline and explore three variants: 1) Baseline U-Net without StyleGAN; 2) Baseline + single scale StyleGAN feature of the smallest size; 3) Baseline + multi-scale StyleGAN features. We also introduce PSNR and SSIM from \cite{pyiqa} as the image quality metrics for evaluating face correction.

\begin{figure}[t]
    \centering
    \includegraphics[width=\linewidth]{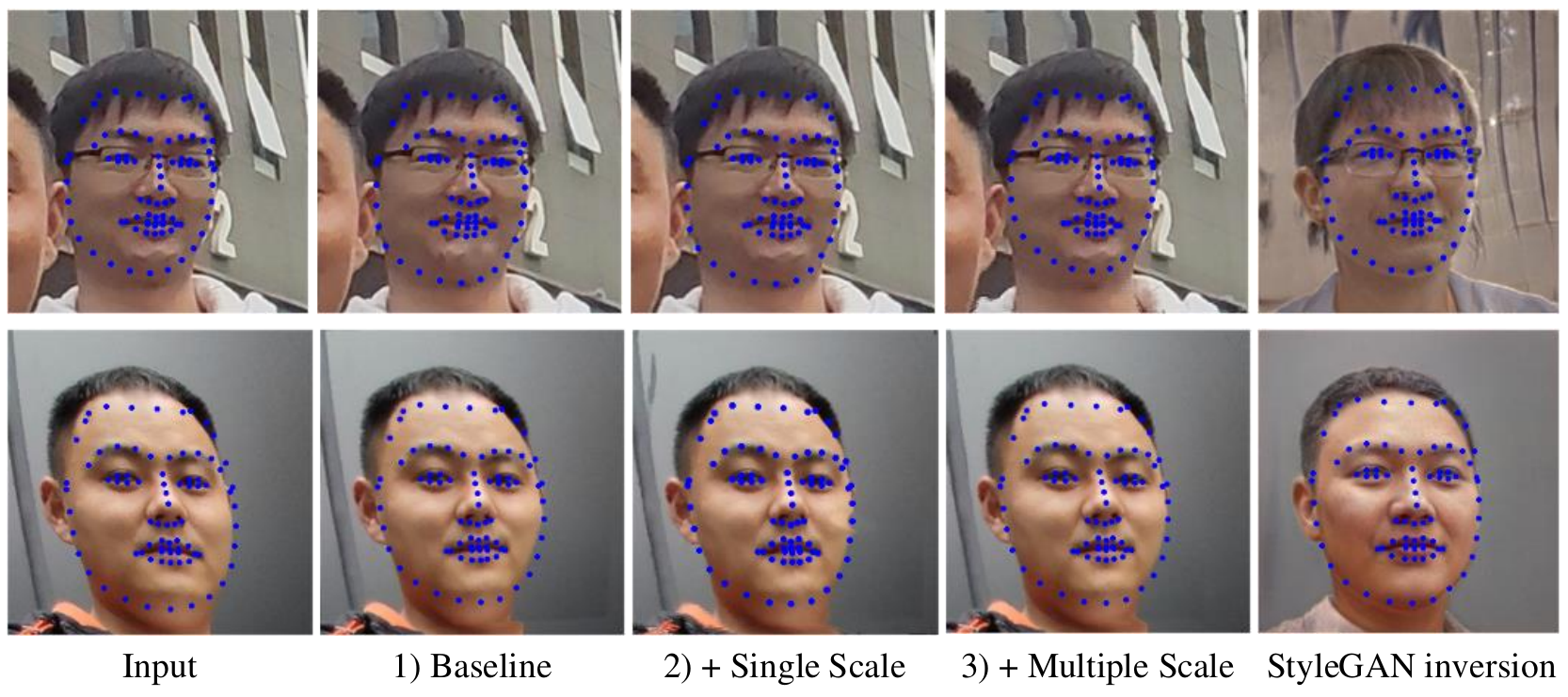}
    \caption{Analyses of multiple scale face prior for face correction.}
    \label{fig:multiscale}
\end{figure}

The quantitative results are summarized in \cref{table:ablation1}. 
Both variants {2)} and~{3)} lead to notable improvements in facial landmarks, with variant {3)} showing a more significant enhancement, achieving the best Landmark Distance of 2.35. This highlights the effectiveness of multi-scale features in enhancing face correction. It is noteworthy that variant {3)} yields better landmark results but slightly lower PSNR/SSIM values. Given that {2)} utilizes high-level scale features, this suggests that the crucial information for face structure correction primarily originates from high-level features, whereas low-level features predominantly influence color and details, thereby improving PSNR/SSIM. \cref{fig:multiscale} shows the qualitative comparisons between different variants as well as the StyleGAN inversion results. Despite the StyleGAN inversion results presenting variations in facial identity, they offer valuable structural guidance for face correction. The results achieved with multiscale features demonstrate improved landmark similarity to StyleGAN inversion compared to those from a single-scale model, highlighting the advantages of multiscale feature fusion in enhancing the accuracy and effectiveness of face correction.

\begin{table}[!t]
\caption{Quantitative comparison of our proposed method and previous works. Results of Shih \etal \cite{shih2019distortion} and Tan \etal \cite{tan2021practical} are directly taken from their papers.}
\label{table:comparison}
\centering
\renewcommand{\arraystretch}{1.1}
  \setlength{\tabcolsep}{2.0pt}
    {
    \begin{tabular}{c|cccc|ccc}
    \toprule
    \multirow{2}{*}{\textbf{Method}} & \multicolumn{2}{c}{\textbf{Note}} & \multicolumn{2}{c|}{\textbf{Vivo}} & \multicolumn{3}{c}{\textbf{All dataset}} \\
    & {LineAcc} & {ShapeAcc} & {LineAcc} & {ShapeAcc} & {LineAcc} & {ShapeAcc} & {Landmark Dis.}\\
    \midrule
    Shih~\cite{shih2019distortion} & - & - & - & - & 66.143 & 97.253 & 6.035 \\
    Tan~\cite{tan2021practical} & 68.683 & 97.115 & 65.148 & 98.363 & 66.784 & 97.490 & - \\
    Zhu~\cite{zhu2022semi} & 66.381 & 97.746 & \textbf{66.572} & 96.076 & 67.209 & 97.500 & 5.840 \\
    Ours & \textbf{69.891} & \textbf{98.697} & 65.459 & \textbf{99.091} & \textbf{67.304} & \textbf{99.012} & \textbf{5.013}\\
    \bottomrule
    \end{tabular}
}
\end{table}

\subsection{Comparison with Other Methods}

\begin{figure}[!t]
  \centering
    \includegraphics[width=.95\linewidth]{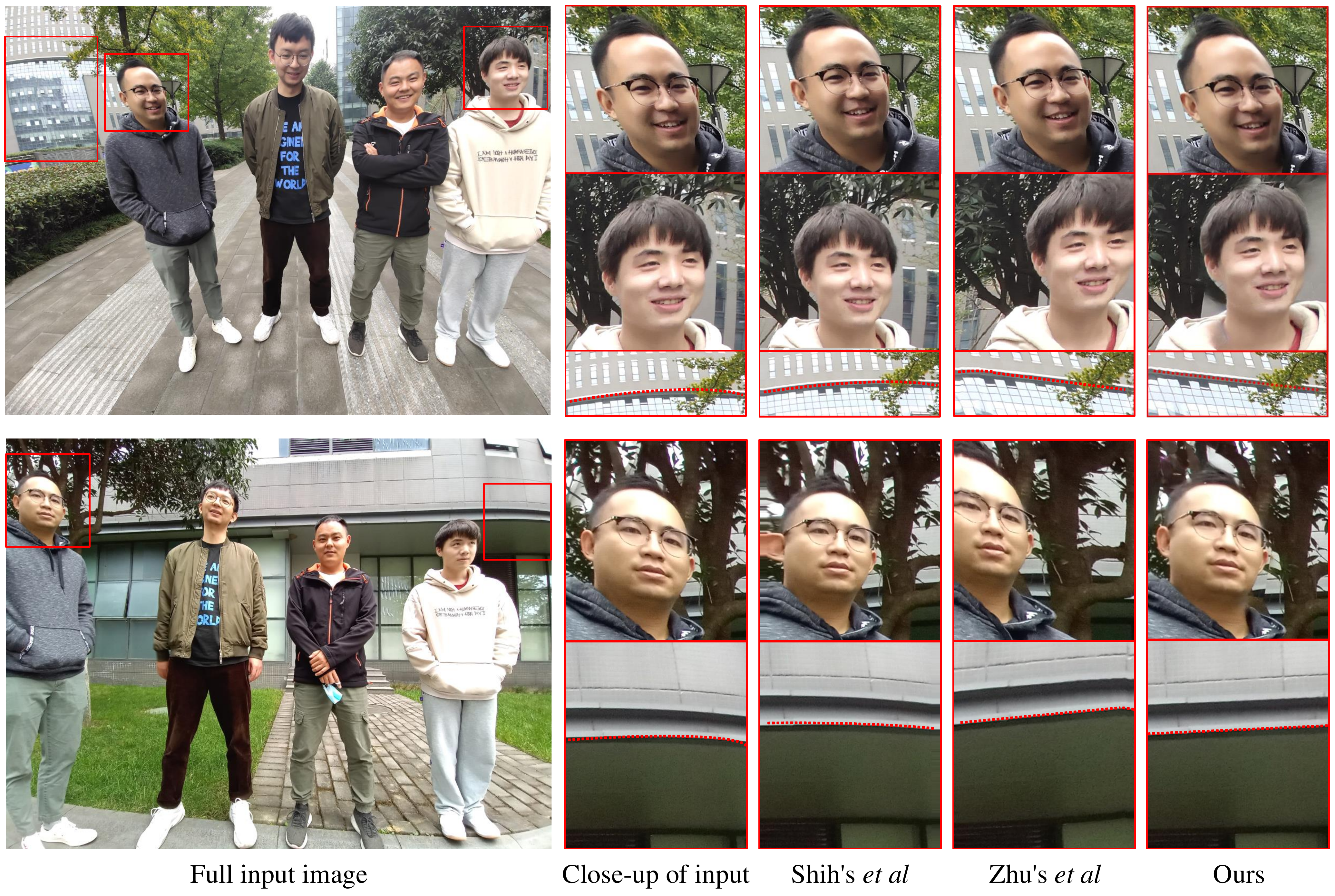}
  \caption{Qualitative results of different correction methods. Our results are significantly better in line correction and yield more natural-looking faces without artifacts.}
  \label{fig:compare}
\end{figure}

%To further show the superiority of our approach, 
We conduct comprehensive comparisons with competing SOTA methods, \ie, Shih \etal \cite{shih2019distortion} and Zhu \etal \cite{zhu2022semi}, and evaluate its effectiveness in both straight line and face restoration tasks.
To ensure a fair comparison, all results are obtained without additional calibration preprocessing.
\cref{table:comparison} provides a detailed comparison of the results achieved by our method against those of existing approaches. 
%Moreover, we report detailed results of various mobile devices (\textsc{Note10} and \textsc{Vivox23}).
The quantitative comparison highlights our method's enhanced ability to preserve straight lines and accurately restore facial features. Notably, our approach significantly reduces the ShapeAcc error by $60\%$, from 2.5 to 1.0. 
To complement the quantitative evaluation, we also conducted visual comparisons, as depicted in \cref{fig:compare}. 
We can observe that our line correction results consistently surpass those of other methods, highlighting the advantage of using a symmetric prior. Furthermore, the faces corrected by our approach appear more natural and are free from artifacts, in contrast to those by Shih \etal in the second row. These comparisons further affirm the efficacy of our method, demonstrating its capability to maintain the original face size while effectively correcting distortions. 
Through these analyses, we have demonstrated the superior performance of our approach in handling both background and facial distortions.

%In summary, our comparative analysis highlights the significant advancements offered by our method over previous state-of-the-art techniques. 
%Through rigorous evaluation and visual demonstrations, we have demonstrated the superior performance and versatility of our approach in handling both straight lines and facial distortions.

\subsection{Limitation}
%Our work aims to incorporate the generative and geometry priors for the wild-angle portrait correction. However, t
The generative face prior may struggle to handle faces with large poses. This can be alleviated by expanding the face diversity for the generative models. Another limitation is the unnatural body region (\eg, obviously longer feet in \cref{fig:compare}). We will solve it by introducing the generative body prior in the future.

\section{Conclusion}
This paper introduces an innovative method to correct wide-angle distortion by leveraging the synergistic capabilities of the generative face prior and geometric symmetry prior. Our approach not only successfully mitigates background distortions through LineCNet, which is trained with symmetric regularization, but also adeptly addresses intricate facial distortions with the incorporation of a StyleGAN prior. The presented visualizations of the correction flow map and facial landmarks highlight the effectiveness of these two priors. Through comprehensive experimentation, our method sets a new benchmark, outperforming existing techniques and showcasing significant visual enhancements.

\section*{Acknowledgements}
This work was supported by the National Natural Science Foundation of China (NSFC) under grant U22B2035.

\bibliographystyle{splncs04}
\bibliography{main}
\end{document}